\documentclass[runningheads]{llncs}
\usepackage{graphicx}

\usepackage{url}
\usepackage{breakurl}
\usepackage[breaklinks]{hyperref}

\usepackage{tikz}
\usepackage{comment}
\usepackage{amsmath,amssymb} 
\usepackage{color}
\usepackage{soul} 
\usepackage{capt-of}

\usepackage[accsupp]{axessibility}  

\usepackage[width=122mm,left=12mm,paperwidth=146mm,height=193mm,top=12mm,paperheight=217mm]{geometry}
\definecolor{my_orange}{rgb}{1.0, 0.55, 0.0}

\newcommand{\method}{Text2LIVE~}

\newcommand{\linkk}[1]{\emph{\textcolor{magenta}{#1}}}
\newcommand{\myparagraph}[1]{\vspace{0.15cm}\noindent{\bf #1}\hspace{0.05cm}} 
\newcommand{\vv}[1]{\mathbf{#1}}
\newcommand\norm[1]{\left\lVert#1\right\rVert}

\begin{document}

\makeatletter
\def\blfootnote{\xdef\@thefnmark{}\@footnotetext}
\makeatother

\newcommand{\afterfigure}{\vspace{-1.3em}}

\pagestyle{headings}
\mainmatter

\title{\mbox{\Large Text2LIVE: Text-Driven Layered Image and Video Editing} \vspace{-0.8cm}} 

\titlerunning{Text2LIVE}
%

\author{Omer Bar-Tal\inst{1*} \and
Dolev Ofri-Amar\inst{1*}\and
Rafail Fridman\inst{1*} \and \\
 Yoni Kasten\inst{2} \and Tali Dekel\inst{1}}
\blfootnote{* Denotes equal contribution.}
%
\authorrunning{O. Bar-Tal, D. Ofri-Amar , R. Fridman et al.}
%
\newcommand{\samelineand}{\qquad \qquad}
\institute{\inst{1} Weizmann Institute of Science \samelineand \inst{2} NVIDIA Research}


\maketitle 
\begin{figure}
    \centering
    \vspace{-0.75cm}
    \includegraphics[width=.99\textwidth]{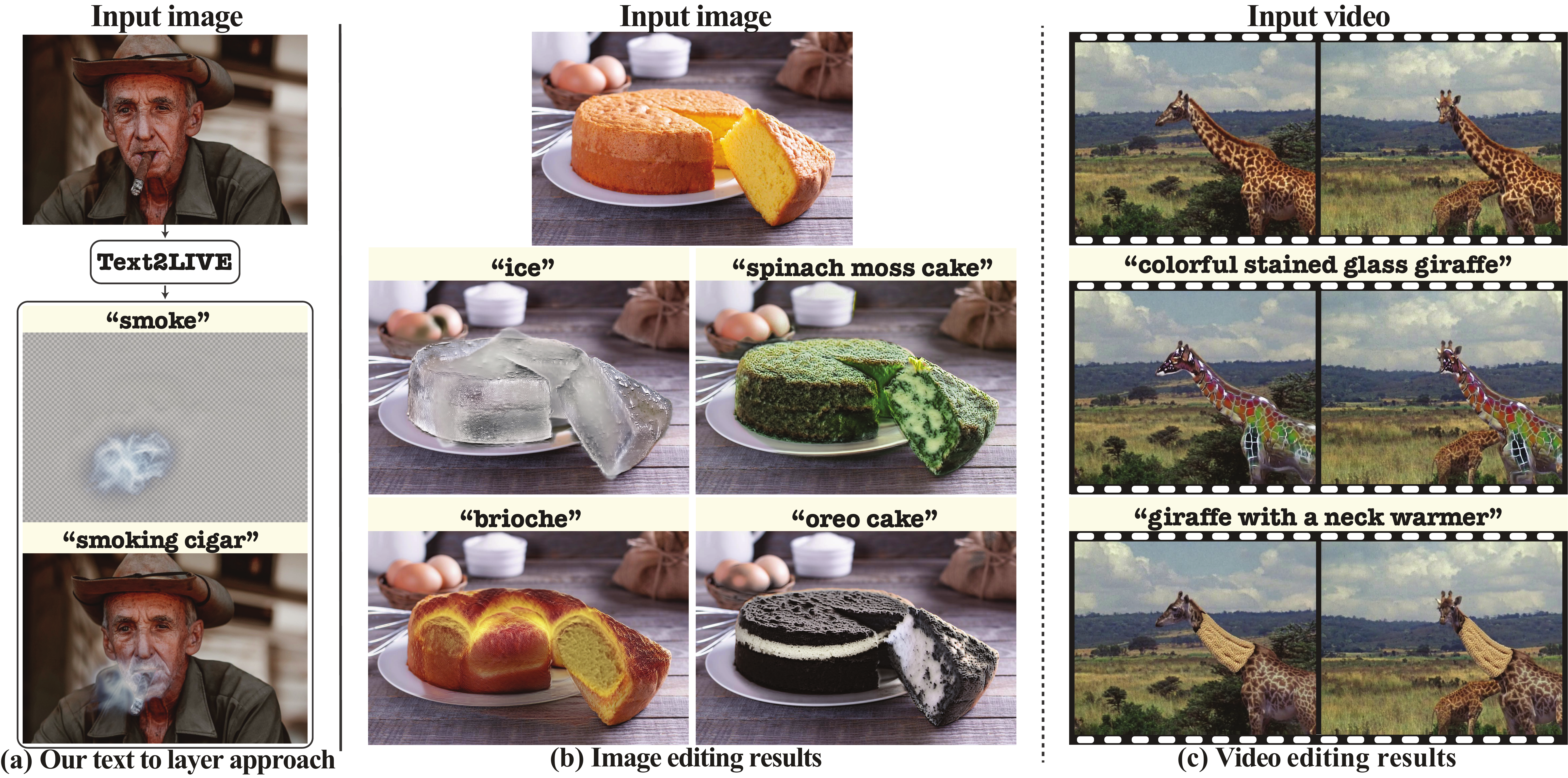} \vspace{-0.4cm}
\caption{\method performs \emph{semantic}, \emph{localized} edits  to  real-world images (b), or videos (c).   Our key idea is to generate an \emph{edit layer}--RGBA image representing the target edit when composited over the original input (a). This allows us to use text to guide not only the final composite, but also the edit layer itself (target text prompts are shown above each image). Our edit layers are synthesized by training a generator on a \emph{single} input,  without relying on user-provided masks or a pre-trained generator.
}
\vspace{-1.1cm}
\label{fig:teaser}
\end{figure}


\begin{abstract}
We present a method for zero-shot, text-driven appearance manipulation in natural images and videos. 
Given an input image or video and a target text prompt, our goal is to edit the appearance of existing objects (e.g., object's texture) or augment the scene with 
visual effects (e.g., smoke, fire) in a semantically meaningful manner. 
We train a generator using an \emph{internal dataset} of training examples, extracted from a single input (image or video and target text prompt), while leveraging an \emph{external} pre-trained CLIP model to establish our losses. Rather than directly generating the edited output, our key idea is to generate an \emph{edit layer} (color+opacity) that is composited over the original input. This allows us to constrain the generation process and maintain high fidelity to the original input via novel text-driven losses that are applied directly to the edit layer. Our method neither relies on a pre-trained generator nor requires user-provided edit masks. We demonstrate localized, semantic edits on high-resolution natural images and videos across a variety of  objects and scenes. Project page: \href{https://text2live.github.io/}{ \linkk{https://text2live.github.io/}}\vspace{-3mm}
 
\end{abstract}

\begin{figure}[t!]
    \centering
    \includegraphics[width=\textwidth]{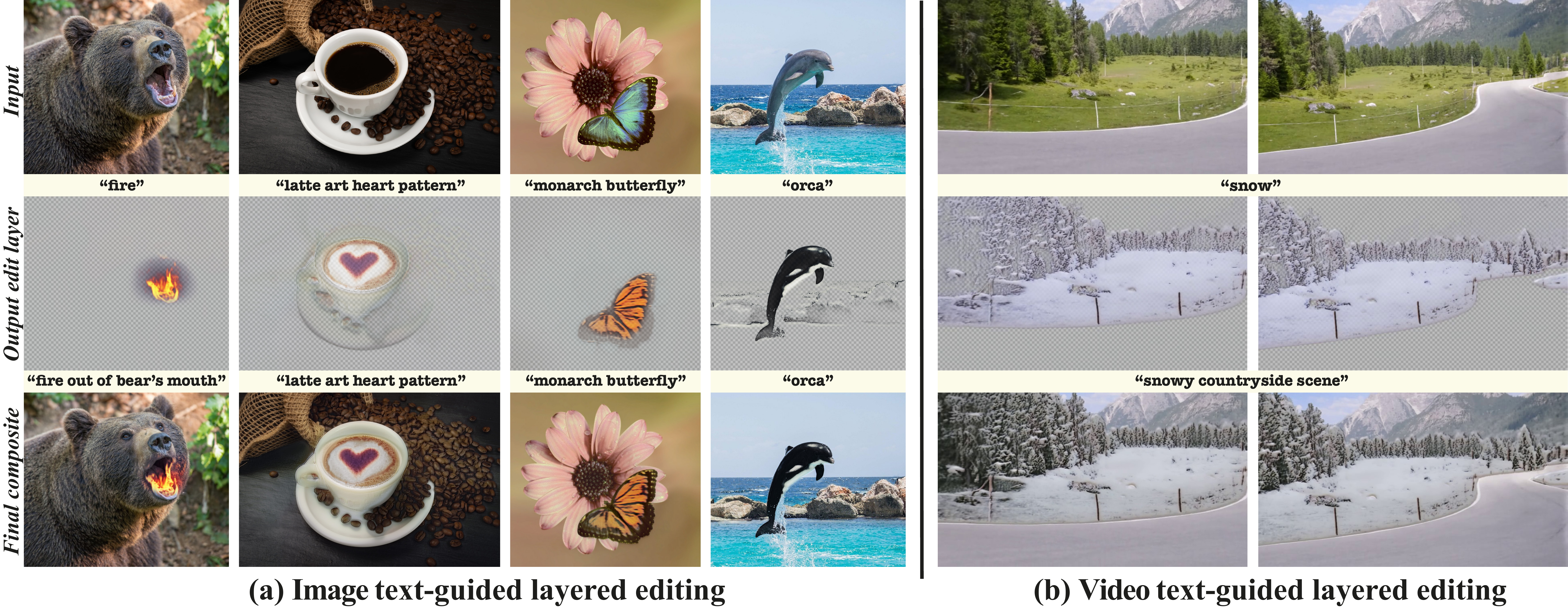} \vspace{-0.5cm}
    \caption{\method generates an edit layer (middle row), which is composited over the original input (bottom row).  The text prompts expressing the target layer and the final composite are shown above each image. Our layered editing facilities a variety of effects including changing objects’ texture or augmenting the scene with complex semi-transparent effects.}
    \label{fig:res_layers}\afterfigure
\end{figure}

\section{Introduction}
Computational methods for manipulating the appearance and style of objects in natural images and videos have seen tremendous progress, facilitating a variety of editing effects to be achieved by novice users. Nevertheless, research in this area has been mostly focused in the Style-Transfer setting where the target appearance is given by a reference image (or domain of images), and the original image is edited in a global manner~\cite{jing2019neural}. Controlling the localization of the edits  typically involves additional input guidance such as segmentation masks. Thus, appearance transfer has been mostly restricted to global artistic stylization or to specific image domains or styles (e.g., faces, day-to-night, summer-to-winter). In this work, we seek to eliminate these requirements and enable more flexible and creative semantic appearance manipulation of real-world images and videos. 

Inspired by the unprecedented power of recent Vision-Language models, we use simple text prompts to express the target edit. 
This allows the user to easily and intuitively specify the target appearance and the object/region to be edited. Specifically, our method enables \emph{local}, \emph{semantic} editing that satisfies a given target text prompt (e.g., Fig.~\ref{fig:teaser} and Fig.~\ref{fig:res_layers}). For example,  given the cake image in Fig.~\ref{fig:teaser}(b), and the target text: ``oreo cake'',  our method automatically locates the cake region and synthesizes realistic, high-quality texture that combines naturally with the original image -- the cream filling and the cookie crumbs ``paint'' the full cake and the sliced piece in a \emph{semantically-aware} manner. As seen, these properties hold across a variety of different edits.

Our framework leverages the representation learned by a Contrastive Language-Image Pretraining (CLIP) model, which has been pre-trained on 400 million text-image examples~\cite{clip}. The richness of the enormous visual and textual space spanned by CLIP has been demonstrated by various recent image editing methods (e.g., \cite{avrahami2021blended,bau2021paint,CLIPDraw,StyleGanNada,StyleCLIP}).
However, the task of editing \emph{existing} objects in \emph{arbitrary, real-world} images remains challenging. Most existing methods combine a pre-trained generator (e.g., a GAN or a Diffusion model) in conjunction with CLIP. With GANs, the domain of images is restricted and requires to invert the input image to the GAN's latent space –-a challenging task by itself~\cite{GANInversionSurvey}. Diffusion models \cite{DDPM,DDIM} overcome these barriers but face an inherent trade-off between satisfying the target edit and maintaining high-fidelity to the original content~\cite{avrahami2021blended}.
Furthermore, it is not straightforward to extend these methods to videos. In this work, we take a different route and propose to \emph{learn a generator from a single input}--image or video and text prompts. 

If no external generative prior is used, how can we steer the generation towards meaningful, high-quality edits? We achieve this via the following two key components: (i) we propose a novel text-guided \emph{layered editing}, i.e.,  rather than directly generating the edited image, we represent the edit via an RGBA layer (color and opacity) that is composited over the input.  This allows us to guide the content and localization of the generated edit via a novel objective function, including text-driven losses applied directly to the edit layer.  For example, as seen in Fig.~\ref{fig:res_layers}, we use text prompts to express not only the final edited image but also a target effect (e.g., fire) represented by the edit layer.  (ii) We train our generator on an \emph{internal dataset} of diverse image-text training examples by applying various augmentations to the input image and text. We demonstrate that our internal learning approach serves as a strong regularization, enabling high quality generation of complex textures and semi-transparent effects.  

We further take our framework to the realm of \emph{text-guided video editing}. Real-world videos often consist of complex object and camera motion, which provide abundant information about the scene. Nevertheless, achieving consistent video editing is difficult and cannot be accomplished na\"ively. We thus propose to decompose  the video into a set of 2D \emph{atlases} using \cite{kasten2021layered}. Each atlas can be treated as a unified 2D image representing either a foreground object or the background throughout the video. This representation significantly simplifies the task of video editing: edits applied to a single 2D atlas are automatically mapped back to the entire video in a consistent manner. We demonstrate how to extend our framework to perform edits in the atlas space while harnessing the rich information readily available in videos. 

In summary, we present the following contributions: 
\begin{itemize}
    \item  An end-to-end text-guided framework for performing localized, semantic edits of existing objects in real-world images.
    \item A novel layered editing approach and objective function that automatically guides the content and localization of the generated edit. 
    \item We demonstrate the effectiveness of internal learning for training a generator on a single input in a zero-shot manner. 
    \item An extension to video which harnesses the richness of information across time, and can perform consistent text-guided editing.  
    \item We demonstrate various edits, ranging from changing objects' texture to generating complex semi-transparent effects, all achieved fully automatically across a wide-range of objects and scenes. 
\end{itemize}
\section{Related Work}

\myparagraph{Text-guided image manipulation and synthesis.} There has been remarkable progress since the use of conditional GANs in both text-guided image generation  \cite{reed2016generative,xu2018attngan,zhang2017stackgan,zhang2018stackgan++}, and editing \cite{dong2017semantic,li2020manigan,nam2018text}. 
ManiGAN~\cite{li2020manigan} proposed a text-conditioned GAN for editing an object's appearance while preserving the image content.  However,  such multi-modal GAN-based methods are restricted to  specific image domains and limited in the expressiveness of the text (e.g., trained on COCO \cite{lin2014microsoft}).  DALL-E~\cite{ramesh2021zero} addresses this by learning a joint image-text distribution over a massive dataset. While achieving remarkable text-to-image generation, DALL-E is not designed for editing existing images.  GLIDE \cite{GLIDE} takes this approach further, supporting both text-to-image generation and inpainting.

Instead of directly training a text-to-image generator, a recent surge of methods leverage a pre-trained generator, and use a pre-trained CLIP~\cite{clip} to guide the generation process by text \cite{bau2021paint,StyleGanNada,liu2021fusedream,StyleCLIP}. StyleCLIP~\cite{StyleCLIP} and StyleGAN-NADA~\cite{StyleGanNada} use a pre-trained StyleGAN2 \cite{stylegan2} for image manipulation, by either controlling the GAN's latent code~\cite{StyleCLIP}, or by fine-tuning the StyleGAN's output domain~\cite{StyleGanNada}. However, editing a real input image using these methods requires first tackling the GAN-inversion challenge~\cite{richardson2021encoding,tov2021designing}. Furthermore, these methods can edit images from a few specific domains, and edit images in a \emph{global} fashion. 
In contrast, we consider a different problem setting -- \emph{localized} edits that can be applied to real-world images spanning a variety of object and scene categories. 

A recent exploratory and artistic trend in the online AI community has demonstrated impressive text-guided image generation. CLIP is used to guide the generation process of a pre-trained generator, e.g., VQ-GAN~\cite{esser2021taming}, or diffusion models~\cite{DDPM,DDIM}. \cite{DiffusionCLIP} takes this approach a step forward by optimizing the diffusion process itself. However, since the generation is \emph{globally} controlled by the diffusion process, this method is not designed to support localized edits that are applied only to selected objects.

To enable region-based editing, user-provided masks are used to control the diffusion process for image inpainting~\cite{avrahami2021blended}. In contrast, our goal is not to generate new objects but rather to manipulate the appearance of existing ones, while preserving the original content. Furthermore, our method is fully automatic and performs the edits directly from the text, without user edit masks.

Several works \cite{CLIPDraw,jain2021dreamfields,CLIPStyler,text2mesh} take a \emph{test-time optimization} approach and leverage CLIP without using a pre-trained generator. For example, CLIPDraw \cite{CLIPDraw} renders a drawing that matches a target text by directly optimizing a set of vector strokes. To prevent adversarial solutions, various augmentations are applied to the output image, all of which are required to align with the target text in CLIP embedding space. CLIPStyler \cite{CLIPStyler} takes a similar approach  for \emph{global} stylization. Our goal is to perform \emph{localized}  edits, which are applied only to specific objects. Furthermore,  CLIPStyler optimizes a CNN  that observes \emph{only} the source image. In contrast, our generator is trained on an  \emph{internal dataset}, extracted from the input image and text. 
We draw inspiration from previous works that show the effectiveness of internal learning in the context of generation ~\cite{SinGAN,InGAN,Splice}.

Other works use CLIP to synthesize \cite{jain2021dreamfields} or edit \cite{text2mesh} a single 3D representation (NeRF or mesh). The unified 3D representation is optimized through a differentiable renderer:  CLIP loss is applied across different 2D rendered viewpoints. Inspired by this approach, we use a similar concept to edit videos. In our case, the ``renderer'' is a layered neural atlas representation of the video~\cite{kasten2021layered}.    

\myparagraph{Consistent Video Editing.} 
Existing approaches for consistent video editing can be roughly divided into: (i) propagation-based methods, which use keyframes \cite{vid_edit_by_example,Texler20-SIG} or optical flow \cite{vid_style_transfer} to propagate edits through the video, and (ii) video layering-based methods, in which a layered representation of the video is estimated and then edited~\cite{kasten2021layered,layer_builder,retiming,omnimatte,unwrap_mosaics}. For example, Lu et al. \cite{retiming,omnimatte} estimate \emph{omnimattes} -- RGBA layers that contain a target subject along with their associated scene effects. Omnimattes facilitate a variety of video effects (e.g., object removal or retiming). However, since the layers are computed independently for each frame, it cannot support consistent propagation of edits across time. Kasten et al. \cite{kasten2021layered} address this challenge by decomposing the video into unified 2D atlas layers (foreground and background). Edits applied to the 2D atlases are automatically mapped back to the video, thus achieving temporal consistency with minimal effort. In our work, we treat a pre-trained neural layered atlas model as a \emph{video renderer} and leverage it for the task of text-guided video editing. 

\begin{figure}[t!]
    \centering
    \includegraphics[width=\textwidth]{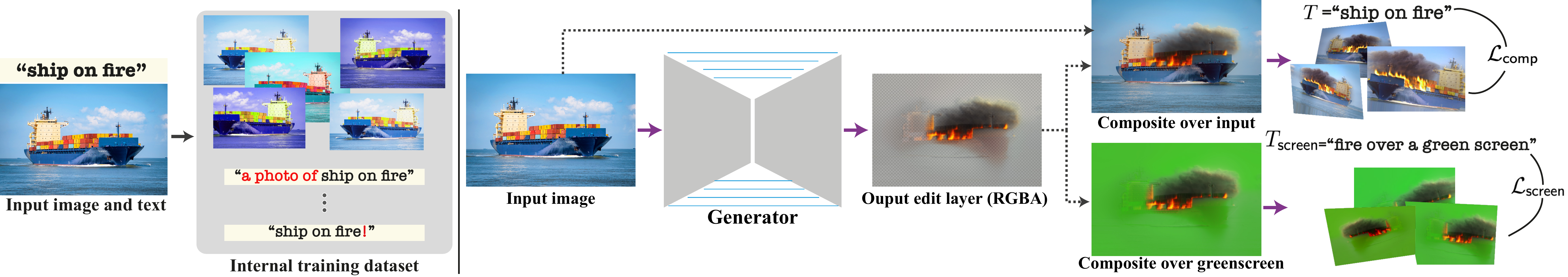}
    \vspace{-0.5cm}
    \caption{{\em Image pipeline.} Our method consists of a generator trained on a single input image and target text prompts. {\em Left:}  an  \emph{internal image-text dataset} of diverse training examples is created by augmenting both image and text (see Sec.~\ref{sec:method_im}). {\em Right:} Our generator takes as input an image and outputs an edit RGBA layer (color+opacity), which is composited over the input to form the final edited image. The generator is trained by minimizing several loss terms that are defined in CLIP space, and include: $\mathcal{L}_{\mathsf{comp}}$, applied to the composite, and $\mathcal{L}_\mathsf{screen}$, applied to the edit layer (when composited over a green background). We apply additional augmentations before CLIP (Sec.~\ref{sec:method_im})} \afterfigure
    \label{fig:pipeline_image}
\end{figure}

\section{Text-Guided Layered Image and Video Editing}
\label{sec:method}
We focus on semantic, localized edits expressed by simple text prompts. Such edits include changing objects' texture or semantically augmenting the scene with complex semi-transparent effects (e.g., smoke, fire).  To this end, we harness the potential of learning a generator from a \emph{single input} image or video while leveraging a pre-trained CLIP model, which is kept fixed and used to establish our losses~\cite{clip}. Our task is ill-posed -- numerous possible edits can satisfy the target text according to CLIP, some of which include noisy or undesired solutions~\cite{CLIPDraw,liu2021fusedream}. Thus, controlling edits' localization and preserving the original content are both pivotal components for achieving high-quality editing results. We tackle these challenges through the following key components:
\begin{enumerate}
    \item {\em Layered editing.} Our generator outputs an RGBA layer that is composited over the input image. This allows us to control the content and spatial extent of the edit via dedicated losses applied directly to the edit layer. 
    \item {\em Explicit content preservation and localization losses.} We devise new losses using the internal spatial features in CLIP space to preserve the original content, and to guide the localization of the edits.  
        \item {\em Internal generative prior.} We construct an internal dataset of examples by applying augmentations to the input image/video and text. These augmented examples are used to train our generator, whose task is to perform text-guided editing on a larger and more diverse set of examples. 
\end{enumerate}

\subsection{Text to Image Edit Layer} \label{sec:method_im}

 As illustrated in Fig.~\ref{fig:pipeline_image}, our framework consists of a generator $G_\theta$ that takes as input a source  image $I_s$ and synthesizes an \emph{edit layer}, $\mathcal{E}=\{C, \alpha\}$, which consists of a color image $C$ and an opacity map $\alpha$. The final edited image $I_o$ is given by compositing the edit layer over $I_s$:
{\small 
\begin{equation}
\label{eq:composite_basic}
    I_o = \alpha \cdot C + (1-\alpha) \cdot I_s
\end{equation}}
Our main goal is to generate $\mathcal{E}$ such that the final composite $I_o$ would comply with a target text prompt $T$. In addition, generating an RGBA layer allows us to use text to further guide the generated content and its localization. To this end, we consider a couple of auxiliary text prompts: $T_{\mathsf{screen}}$  which expresses the target \emph{edit layer}, when composited over a green background, and $T_\mathsf{ROI}$ which specifies a region-of-interest in the source image, and is used to initialize the localization of the edit.  For example, in the \emph{Bear} edit in Fig.~\ref{fig:res_layers}, $T=$\emph{``fire out of the bear's mouth''},  $T_\mathsf{screen}=$\emph{``fire over a green screen''}, and $T_\mathsf{ROI}=$\emph{``mouth''}. We next describe in detail how these are used in our objective function. 

\myparagraph{Objective function.} 
Our novel objective function incorporates three main loss terms, all defined in CLIP's feature space: (i) $\mathcal{L}_{\mathsf{comp}}$, which is the driving loss and encourages ${I_o}$ to conform with $T$, (ii) $\mathcal{L}_{\mathsf{screen}}$, which serves as a direct supervision on the edit layer, and (iii) $\mathcal{L}_{\mathsf{structure}}$, a structure preservation loss w.r.t. ${I_s}$. Additionally, a regularization term $\mathcal{L}_{\mathsf{reg}}$ is used for controlling the extent of the edit by encouraging sparse alpha matte $\alpha$. Formally, 

{\small \begin{equation} \label{eq:total_loss}
    \mathcal{L}_{\mathsf{Text2LIVE}} =   \mathcal{L}_{\mathsf{comp}} + \lambda_g \mathcal{L}_{\mathsf{screen}} + \lambda_s \mathcal{L}_{\mathsf{structure}} + \lambda_r \mathcal{L}_{\mathsf{reg}},
\end{equation}}
where $\lambda_g$, $\lambda_s$, and $\lambda_r$ control the relative weights between the terms, and are fixed throughout all our experiments (see Appendix~\ref{sec:training_details}).

\myparagraph{Composition loss.}  $\mathcal{L}_{\mathsf{comp}}$ reflects our primary objective of generating an image that matches the target text prompt and is given by a combination of a \emph{cosine distance} loss and a \emph{directional} loss~\cite{StyleCLIP}: 
{\small
\begin{equation} \label{eq:6}
    \mathcal{L}_{\mathsf{comp}} = \mathcal{L}_\mathsf{cos}\left(I_o, T\right) +
    \mathcal{L}_{\mathsf{dir}}(I_s, I_o, T_{\mathsf{ROI}}, T), 
\end{equation}}
where {\small $\mathcal{L}_\mathsf{cos}=\mathcal{D}_\text{cos}\left(E_\text{im}(I_o),E_\text{txt}(T)\right)$} is the cosine distance between the  CLIP embeddings for $I_o$ and $T$. Here, {\small $E_\text{im}$, $E_\text{txt}$} denote CLIP's image and text encoders, respectively. The second term controls the direction of edit in CLIP space~\cite{StyleGanNada,StyleCLIP} and is given by: {\small $\mathcal{L}_{\mathsf{dir}}=\mathcal{D}_\text{cos}(E_\text{im}(I_o)\!-\!E_\text{im}(I_s), E_\text{txt}(T)-E_\text{txt}(T_{\mathsf{ROI}}))$} .

Similar to most CLIP-based editing methods, we first augment each image to get several different views and calculate the CLIP losses w.r.t. each of them separately, as in~\cite{avrahami2021blended}. This holds for all our CLIP-based losses. See Appendix~\ref{sec:internal-dataset} for details.  

\myparagraph{Screen loss.} The term $\mathcal{L}_{\mathsf{screen}}$ serves as a direct text supervision on the generated edit layer $\mathcal{E}$. We draw inspiration from chroma keying~\cite{brinkmann2008art}--a well-known  technique by which a solid background (often green) is replaced by an image in a post-process. Chroma keying is extensively used in image and video post-production, and there is high prevalence of online images depicting various visual elements over a green background. We thus composite the edit layer over a green background $I_{\mathsf{green}}$ and encourage it to match the text-template $T_{\mathsf{screen}}\!:=\!``\,\{~\}$\emph{ over a green screen"}, (Fig.~\ref{fig:pipeline_image}):
{\small
\begin{equation} \label{eq:6}
    \mathcal{L}_{\mathsf{screen}} =  \mathcal{L}_\mathsf{cos}\left(I_{\mathsf{screen}}, T_{\mathsf{screen}}\right) 
\end{equation}}
where $I_{\mathsf{screen}}= \alpha \cdot C + (1-\alpha)\cdot I_{\mathsf{green}}$. 

A nice property of this loss is that it allows intuitive supervision on a desired \emph{effect}. For example, when generating semi-transparent effects, e.g., \emph{Bear} in Fig.~\ref{fig:res_layers}, we can use this loss to focus on the fire regardless of the image content by using $T_{\mathsf{screen}}=$``fire over a green screen''. Unless specified otherwise, we plug in $T$ to our screen text template in all our experiments. Similar to the composition loss, we first apply augmentations on the images before feeding to CLIP. 

\myparagraph{Structure loss.} We want to allow substantial texture and appearance changes while preserving the objects' original spatial layout, shape, and perceived semantics. While various perceptual content losses have been proposed in the context of style transfer, most of them use features extracted from a pre-trained VGG model. Instead, we define our loss in CLIP feature space. This allows us to impose additional constraints to the resulting internal CLIP representation of $I_o$.  Inspired by classical and recent works \cite{STROTSS,shechtman2007localselfsim,Splice}, we adopt the \emph{self-similarity} measure. Specifically, we feed an image into CLIP's ViT encoder and extract its $K$ spatial tokens from the deepest layer. The self-similarity matrix, denoted by $S(I)\in \mathbb{R}^{K\times K}$, is used as structure representation. Each matrix element $S(I)_{ij}$ is defined by:  
{\small 
\begin{equation}
S(I)_{ij} = 1-\mathcal{D}_\text{cos}\left(\vv{t}^{i}(I), \vv{t}^{j}(I)\right)
\label{eq:selfsim}
\end{equation}  }
where $\vv{t}_i(I)\in\mathbb{R}^{768}$ is the $i^\text{th}$ token of image $I$. 

The term $\mathcal{L}_{\mathsf{structure}}$ is defined as the Frobenius norm distance between the self-similarity matrices of  $I_s$, and $I_o$:
{\small \begin{equation} \label{eq:5}
    \mathcal{L}_{\mathsf{structure}} = \left\|S(I_s) - S(I_o) \right\|_F
\end{equation}}

\myparagraph{Sparsity regularization.} To control the spatial extent of the edit, we encourage the output opacity map to be sparse. We follow \cite{retiming,omnimatte} and define the sparsity loss term as a combination of $L_1$- and $L_0$-approximation regularization terms:  
{\small 
\begin{equation}
\mathcal{L}_{\mathsf{reg}} = \gamma \norm{\alpha}_1+\Psi_0(\alpha)
\label{eq:sparsity}
\end{equation}} 
where $\Psi_0(x) \equiv 2 \text{Sigmoid}(5x)-1 $ is  a smooth $L_0$ approximation that penalizes non zero elements. We fix $\gamma$ in all our experiments.  

\myparagraph{Bootstrapping.} To achieve accurate localized effects without user-provided edit mask, we apply a text-driven \emph{relevancy} loss to initialize our opacity map.  Specifically, we use Chefer et al. \cite{Chefer_2021_ICCV} to automatically estimate a \emph{relevancy map}\footnote{\cite{Chefer_2021_ICCV} can only work with $224\times 224$ images, so we resize both $I_s$ and $\alpha$ to $224\times 224$ before applying the loss of \eqref{eq:bootstrapping}} $R(I_s)\in [0,1]^{224 \times 224}$  which roughly highlights the image regions that are most relevant to a given text $T_\mathsf{ROI}$. We use the relevancy map to initialize  $\alpha$ by minimizing: 
{\small 
\begin{equation}
\mathcal{L}_{\mathsf{init}} = \text{MSE}\left(R(I_s), \alpha\right)
\label{eq:bootstrapping}
\end{equation}}  
Note that the relevancy maps are noisy, and only provide a rough estimation for the region of interest (Fig.~\ref{fig:bootstrapping}(c)). Thus, we anneal this loss during training (see implementation details in Appendix~\ref{sec:training_details}). By training on diverse internal examples along with the rest of our losses, our framework dramatically refines this rough initialization, and produces accurate and clean opacity (Fig.~\ref{fig:bootstrapping}(d)). 

\myparagraph{Training data.} Our generator is trained from scratch for each input $(I_s, T)$  using an \emph{internal dataset} of diverse image-text training examples $\{(I^i_s, T^i)\}_{i=1}^N$ that are derived from the input (Fig.~\ref{fig:pipeline_image} left). Specifically, each training example $(I^i_s, T^i)$  is generated by randomly applying a set of augmentations to $I_s$ and to $T$. The image augmentations include global crops, color jittering, and flip, while text augmentations are randomly sampled from a predefined text template  (e.g., \emph{``a photo of''$+T$}); see Appendix~\ref{sec:internal-dataset} for details. 
The vast space of all combinations between these augmentations provides us with a rich and diverse dataset for training. The task is now to learn \emph{one} mapping function $G_\theta$ for the \emph{entire dataset}, which poses a strong regularization on the task. Specifically, for each individual example, $G_\theta$ has to generate a plausible edit layer $\mathcal{E}^i$ from $I_s^i$ such that the composited image is well described by $T^i$. We demonstrate the effectiveness of our \emph{internal learning} approach compared to the test-time optimization approach in Sec.~\ref{sec:results}. 

\begin{figure}[t!]
    \centering
    \includegraphics[width=\textwidth]{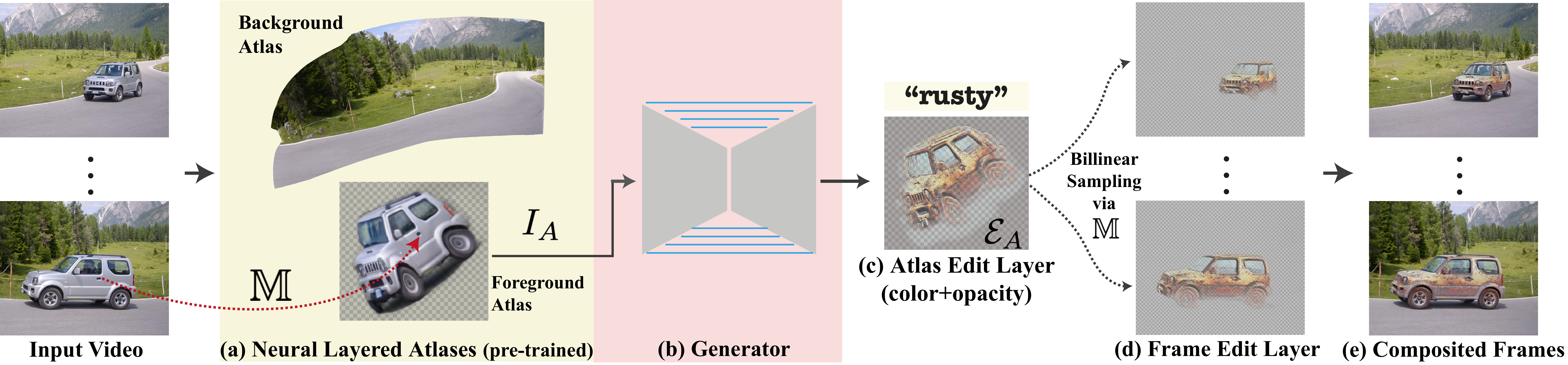} \vspace{-0.5cm}
    \caption{ \emph{Video pipeline.}  (a) a pre-trained and fixed \emph{layered neural atlas} model~\protect\cite{kasten2021layered} is used as a ``video renderer'', which consists of: a set of 2D atlases,  mapping functions from pixels to the atlases (and per-pixel fg/bg opacity values). (b) Our framework trains a generator  that takes a chosen (discretized) atlas $I_A$ as input and a target text prompt (e.g., ``rusty car''), and outputs (c) an atlas edit layer $\mathcal{E}_A$. (d) The edited atlas is rendered back to frames using the pre-trained mapping network $\mathbb{M}$, and then  (e) composited over the original video.}
    \label{fig:pipeline_video}\afterfigure
\end{figure}

\subsection{Text to Video Edit Layer}
\label{sec:method_vid}
A natural question is whether our image framework can be applied to videos. The key additional challenge is achieving a temporally consistent result.  Na\"ively applying our image framework on each frame independently yields unsatisfactory jittery results~(see Sec.~\ref{sec:results}). To enforce temporal consistency, we utilize the Neural Layered Atlases (NLA) method~\cite{kasten2021layered},  as illustrated in Fig.~\ref{fig:pipeline_video}(a). We next provide a brief review of NLA and discuss in detail how our extension to videos.

\myparagraph{Preliminary: Neural Layered Atlases.} NLA provides a unified 2D parameterization of a video: the  video is decomposed into a set of 2D atlases, each can be treated as a  2D image, representing either one foreground object or the  background throughout the entire video. An example of foreground and background atlases are shown in Fig.~\ref{fig:pipeline_video}. For each video location $p=(x,y,t)$, NLA computes a corresponding 2D location (UV) in each atlas, and a foreground opacity value. This allows to reconstruct the original video from the set atlases.  NLA comprises of several Multi-Layered Perceptrons (MLPs), representing the atlases, the mappings from pixels to atlases and their opacity. More specifically, each video location $p$ is first fed into two mapping networks, $\mathbb{M}_b$ and $\mathbb{M}_f$:
{\small \begin{equation}
     \mathbb{M}_b(p) = (u_b^p,v_b^p), \ \ \ \   \mathbb{M}_f(p) = (u_f^p,v_f^p)
\end{equation}}
where $(u_*^p, v_*^p)$ are the 2D coordinates in the background/foreground atlas space. Each pixel is also fed to an MLP that predicts the opacity value of the foreground in each position. 
The predicted UV coordinates are then fed into an atlas network $\mathbb{A}$,  which outputs the  RGB colors in each  location.  Thus, the original RGB value of $p$ can be reconstructed by mapping $p$ to the atlases, extracting the corresponding atlas colors, and blending them  according to the predicted opacity. We refer the reader to \cite{kasten2021layered} for full details. 

Importantly, NLA enables consistent video editing: the continuous atlas (foreground or background) is first discretized to a fixed resolution image (e.g., 1000$\times$1000 px). 
The user can directly edit the discretized atlas using image editing tools (e.g., Photoshop). The atlas edit is then
mapped back to the video, and blended with the original  frames, using the predicted UV mappings and foreground opacity. 
In this work,  we are interested in generating atlas edits in a \emph{fully automatic} manner, solely guided by text.

\myparagraph{Text to Atlas Edit Layer.} Our video framework leverages  NLA as a ``video renderer'', as illustrated in Fig.~\ref{fig:pipeline_video}.  Specifically, given a pre-trained and fixed NLA model for a video, our goal is to generate a 2D \emph{atlas edit layer}, either for the background or foreground, such that when mapped back to the video, each of the rendered frames would comply with the target text.  

Similar to the image framework, we train a generator $G_\theta$ that takes a 2D atlas as input and generates an atlas edit layer $\mathcal{E}_A=\{C_A, \alpha_A\}$. Note that since $G_\theta$ is a CNN, we work with a discretized atlas, denoted as $I_A$.  The pre-trained UV mapping, denoted by $\mathbb{M}$, is used to bilinearly sample $\mathcal{E}_A$ to map it to each frame:
{\small \begin{equation}
  \label{eq:atlas_sampling}
  \mathcal{E}_\text{t} = \text{Sampler}(\mathcal{E}_A, \mathcal{S})
\end{equation}}
where $\mathcal{S}=\{\mathbb{M}(p) ~ |~ p=(\cdot,\cdot,t)\}$ is the set of UV coordinates that correspond to frame $t$. The final edited video is obtained by blending $\mathcal{E}_t$ with the original frames, following the same process as done in \cite{kasten2021layered}.

\begin{figure}[t!]
    \centering
    \includegraphics[width=\textwidth]{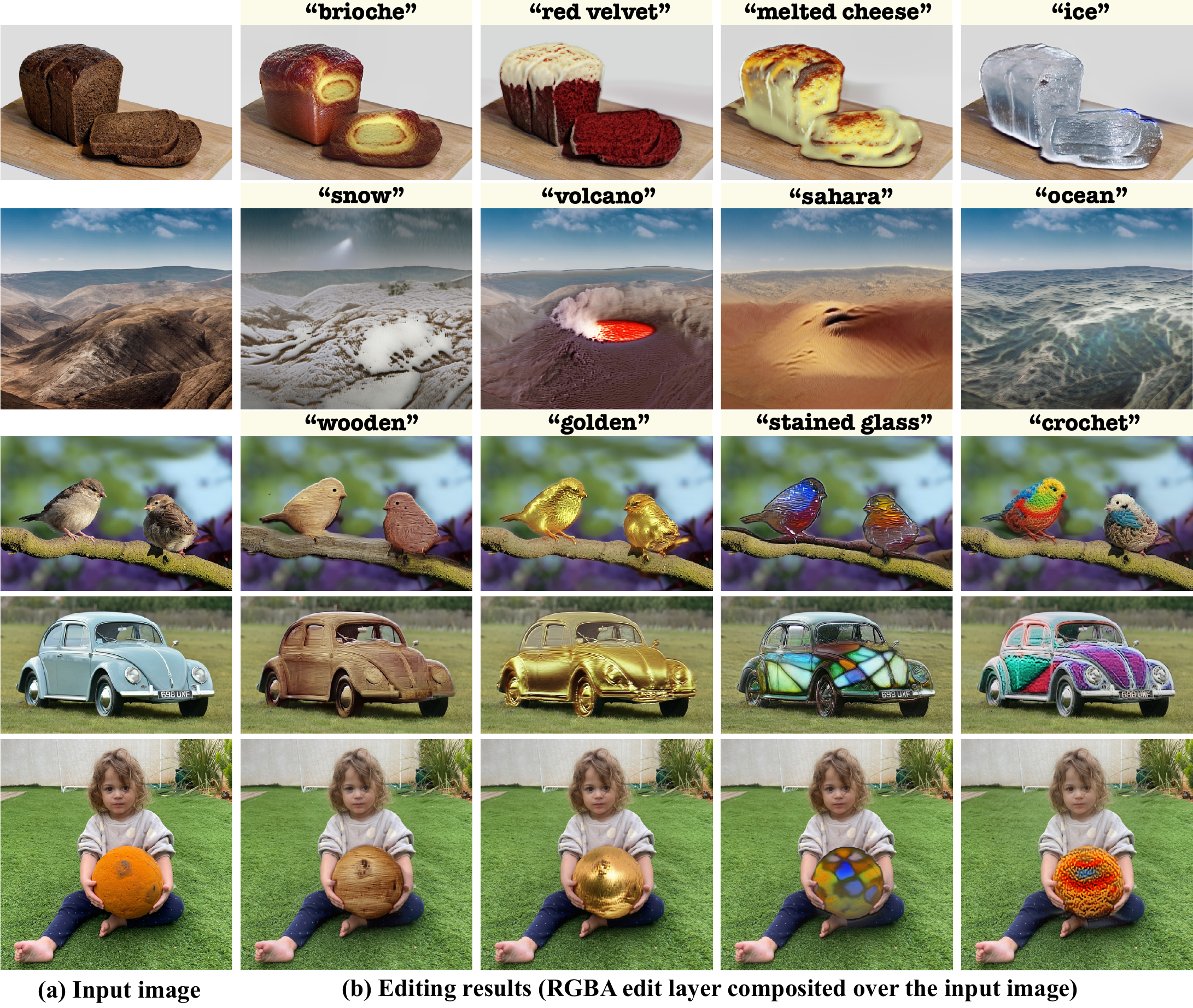}\vspace{-0.3cm}
    \caption{{\em \method image results.} Across rows: different images, across columns: different target edits. All results are produced fully automatically w/o any input masks. 
    }\afterfigure
    \label{fig:object_paint_res}
\end{figure}

\begin{figure}[t!]
    \centering
    \includegraphics[width=\textwidth]{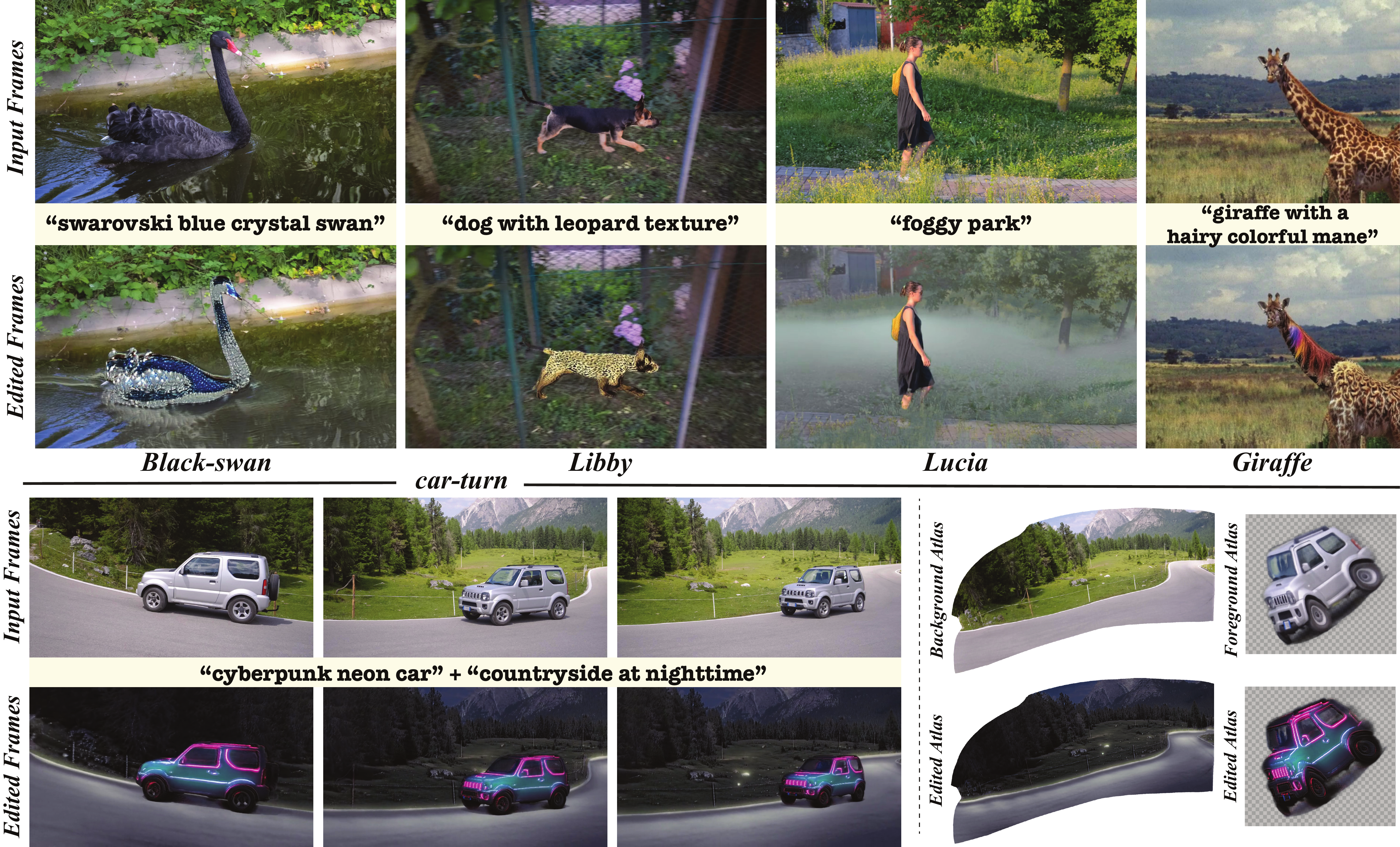} \vspace{-0.8cm}
    \caption{{\em \method video results.} A representative frame from the original and edited videos are shown for each example, along with the target text prompt. In  \emph{car-turn}, both foreground and background atlases are edited sequentially (see Sec.~\ref{sec:results}). The original and edited atlases are shown on the right. Full video results are included in the SM.}
    \label{fig:results_video}
\end{figure}

\myparagraph{Training.}  A straightforward approach for training $G_\theta$ is to treat $I_A$ as an image and plug it into our image framework (Sec.~\ref{sec:method_im}). This approach will result in a temporally consistent result, yet it has two main drawbacks: (i) the atlas often non-uniformly distorts the original structures (see Fig. \ref{fig:pipeline_video}), which may lead to low-quality edits , (ii) solely using the atlas, while ignoring the video frames, disregards the abundant, diverse information available in the video such as different viewpoints, or non-rigid object deformations, which can serve as ``natural augmentations'' to our generator.  We overcome these drawbacks by  mapping the atlas edit back to the video and applying our losses  on the resulting edited \emph{frames}. Similar to the image case, we use the same objective function (Eq.~\ref{eq:total_loss}), and construct an internal dataset directly from the atlas for training. 

More specifically, a training example is constructed by first extracting a crop from $I_A$. To ensure we sample informative atlas regions, we first randomly crop a video segment in both space and time, and then map it to a corresponding atlas crop $I_{Ac}$ using $\mathbb{M}$ (see Appendix~\ref{sec:video_details} for full technical details). We then apply additional augmentations to $I_{Ac}$ and feed it into the generator, resulting in an edit layer $\mathcal{E}_{Ac}=G_\theta(I_{Ac})$. We then map $\mathcal{E}_{Ac}$ and $I_{Ac}$ back to the video, resulting in frame edit layer $\mathcal{E}_t$, and a reconstructed foreground/background crop $I_t$. This is done by bilinearly sampling $\mathcal{E}_{Ac}$ and $I_{Ac}$ using  Eq.~\eqref{eq:atlas_sampling}, with $\mathcal{S}$ as the set of UV coordinates corresponding to the frame crop.  Finally, we apply $\mathcal{L}_{\mathsf{Text2LIVE}}$ from Eq.~\ref{eq:total_loss}, where $I_s\!=I_{t}$ and $\mathcal{E}=\mathcal{E}_{t}$.

\begin{figure}[t!]
    \centering
    \includegraphics[width=\textwidth]{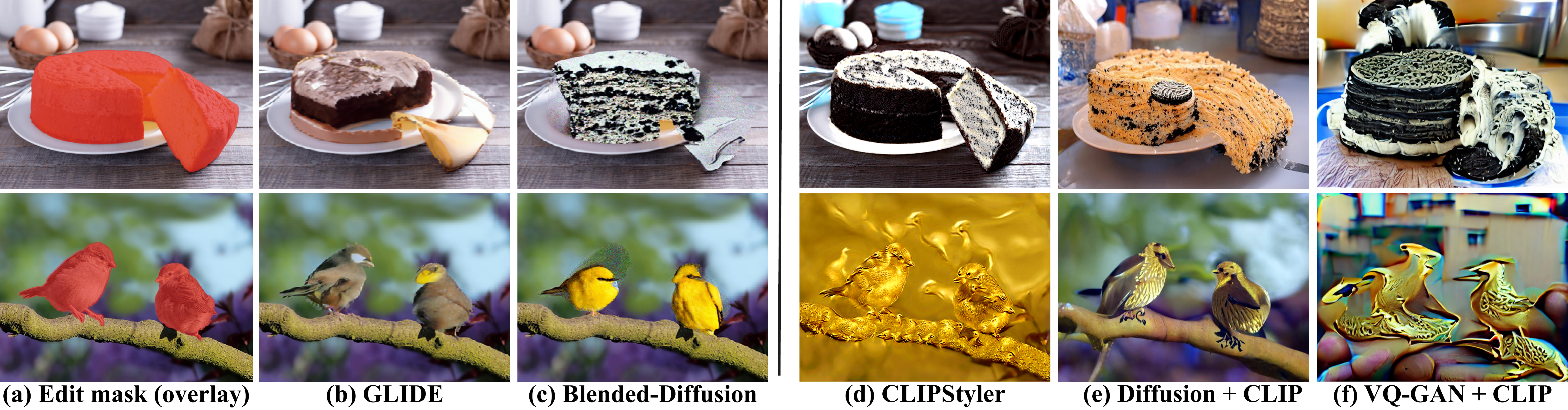}\vspace{-0.4cm}
    \caption{\emph{Comparison to baselines.} A couple of inputs are plugged into different image manipulation methods: \emph{cake} image, shown in Fig.~\ref{fig:teaser}, using ``oreo cake''; and \emph{birds}, shown in Fig.~\ref{fig:object_paint_res},  using ``golden birds''.  (a)   manually created masks (shown in red over the input) are provided to  (b-c) the inpainting methods as additional inputs, while the rest of the methods are mask-free. Our results are shown in Fig.~\ref{fig:teaser}, and Fig.~\ref{fig:object_paint_res}.}\afterfigure
    \label{fig:comp_illustration}
\end{figure}

\section{Results} \label{sec:results}

\subsection{Qualitative evaluation}
We tested our method across various real-world, high-resolution images and videos. The image set contains 35 images collected from the web, spanning various object categories, including animals, food, landscapes and others. The video set contains seven videos from DAVIS dataset \cite{pont20172017}. We applied our method using various target edits, ranging from text prompts that describe the texture/materials of specific objects, to edits that express complex scene effects such as smoke, fire, or clouds. Sample examples for the inputs along with our results can be seen in Fig. \ref{fig:teaser}, Fig.~\ref{fig:res_layers}, and Fig. \ref{fig:object_paint_res} for images, and Fig. \ref{fig:results_video} for videos. The full set of examples and results are included in the Supplementary Materials (SM). As can be seen, in all examples, our method successfully generates photorealistic textures that are ``painted'' over the target objects in a semantically aware manner. For example, in \emph{red velvet} edit (first row in Fig. \ref{fig:object_paint_res}), the frosting is naturally placed on the top. In \emph{car-turn} example (Fig.~\ref{fig:results_video}), the neon lights nicely follow the car's framing. In all examples, the edits are accurately localized,  even under partial occlusions, multiple objects (last row and third row of Fig. \ref{fig:object_paint_res}) and complex scene composition (the dog in Fig. \ref{fig:res_layers}).  Our method successfully augments the input scene with complex semi-transparent effects without changing irrelevant content in the image (see Fig. \ref{fig:teaser}).

\begin{figure}[t!]
    \centering
    \includegraphics[width=\textwidth]{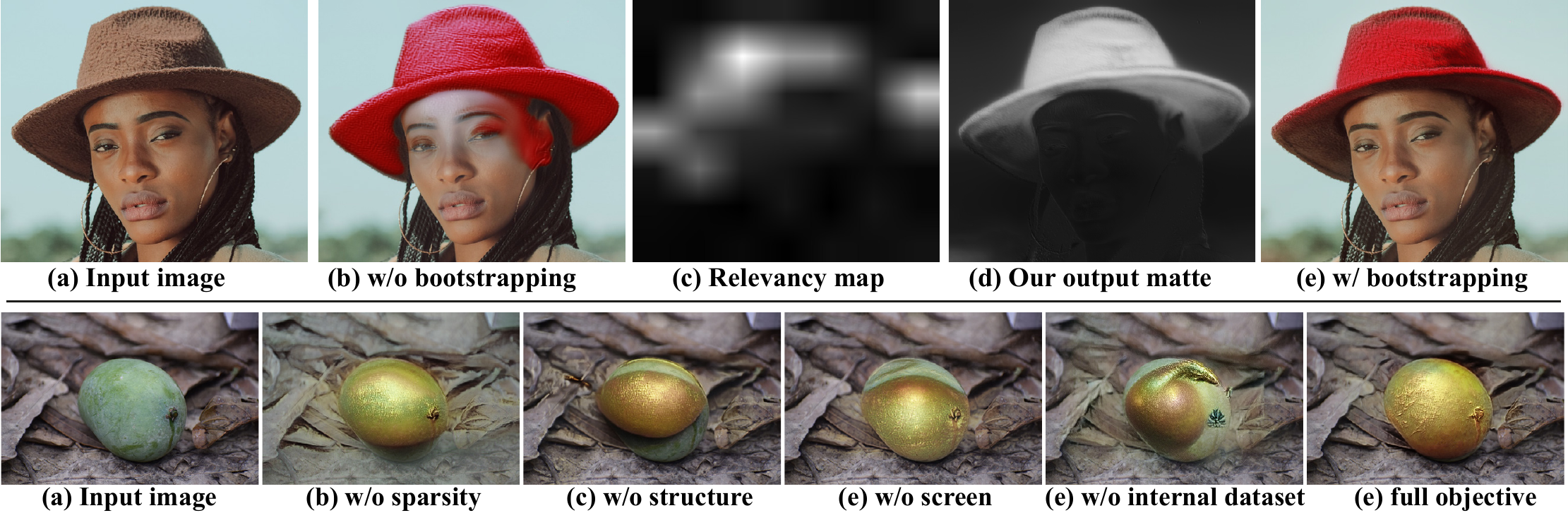} \vspace{-0.8cm}
    \caption{\emph{Top:} We illustrate the effect of our relevancy-based bootstrapping for image (a) using ``red hat'' as the target edit. (b) w/o bootstrapping our edited image suffers from color bleeding. When initializing our alpha-matte to capture the hat ($T_\mathsf{ROI}=$``hat''), an accurate matting is achieved (d-e). Notably, the raw relevancy map provides very rough supervision (c); during training, our method dramatically refines it (d). \emph{Bottom:} We ablate each of our loss terms and the effect of internal learning (``mango'' to ``golden mango"). See Sec.~\ref{sec:ablation}.} \afterfigure
    \label{fig:bootstrapping}
\end{figure}

\subsection{Comparison to Prior Work}
To the best of our knowledge, there is no existing method tailored for solving our task: text-driven \emph{semantic}, \emph{localized} editing of \emph{existing} objects in \emph{real-world} images and videos. We illustrate the key differences between our method and several prominent text-driven image editing methods. We consider those that can be applied to a similar setting to ours: editing real-world images that are not restricted to specific domains. Inpainting methods: Blended-Diffusion \cite{avrahami2021blended} and GLIDE~\cite{GLIDE}, both require user-provided editing mask. CLIPStyler, which performs image stylization, and Diffusion+CLIP \cite{diffClipNotebook}, and  VQ-GAN+CLIP \cite{vqganClip}: two  baselines that combine CLIP with either a pre-trained VQ-GAN or a Diffusion model. In the SM, we also include additional qualitative comparison to the~StyleGAN text-guided editing methods \cite{StyleCLIP,StyleGanNada}.

Fig. \ref{fig:comp_illustration} shows representative results, and the rest are included in the SM. 
As can be seen, none of these methods are designed for our task. The inpainting methods (b-c), even when supplied with tight edit masks, generate new content in the masked region rather than changing the texture of the existing one. CLIPStyler modifies the image in a \emph{global} artistic manner, rather than performing \emph{local} semantic editing (e.g., the background in both examples is entirely changed, regardless of the image content). For the baselines (d-f), Diffusion+CLIP \cite{diffClipNotebook} can often synthesize high-quality images, but with either low-fidelity to the target text (e), or with low-fidelity to the input image content (see many examples in SM). VQ-GAN+CLIP \cite{vqganClip} fails to maintain fidelity to the input image and produces non-realistic images (f). Our method automatically locates the cake region and generates high-quality texture that naturally combines with the original content.  
\subsection{Quantitative evaluation}
\myparagraph{Comparison to image baselines.}
We  conduct an extensive human perceptual evaluation on Amazon Mechanical Turk (AMT). We adopt the Two-alternative Forced Choice (2AFC) protocol suggested in \cite{STROTSS,swapA}. Participants are shown a reference image and a target editing prompt, along with two alternatives: our result and another baseline result. We consider from the above baselines those not requiring user-masks. The participants are asked: \emph{``Which image better shows objects in the reference image edited according to the text"}. We perform the survey using a total of 82 image-text combinations. We collected 12,450 user judgments w.r.t. prominent text-guided image editing methods. Table \ref{table:AMT} reports the percentage of votes in our favor. As seen, our method outperforms all baselines by a large margin, including those using a strong generative prior.

\myparagraph{Comparison to video baselines.}
We quantify the effectiveness of our key design choices for the video-editing by comparing our video method against: (i) \emph{Atlas Baseline:} feeding the discretized 2D Atlas to our single-image method (Sec. \ref{sec:method_im}), and using the same inference pipeline  illustrated in Fig. \ref{fig:pipeline_video} to map the edited atlas back to frames. (ii) \emph{Frames Baseline:} treating all video frames as part of a single   \emph{internal dataset}, used to train our generator; at inference, we apply the trained generator independently to each frame.

We conduct a human perceptual evaluation in which we provide participants a target editing prompt and two video alternatives: our result and a baseline. The participants are asked \emph{``Choose the video that has better quality and better represents the text"}. We collected 2,400 user judgments over 19 video-text combinations and report the percentage of votes in favor of the complete model in table \ref{table:AMT}. We first note that the \emph{Frames baseline} produces temporally inconsistent edits. As expected, the \emph{Atlas baseline} produces temporally consistent results. However, it struggles to generate high-quality textures and often produces blurry results. These observations support our hypotheses  mentioned in Sec.~\ref{sec:method_vid}. We refer the reader to the SM for visual comparisons.
\begin{table}[t!]
\begin{center}
\begin{tabular}{ccc|cc}
\hline
 \multicolumn{3}{c|}{Image baselines} &  \multicolumn{2}{c}{Video baselines} \\
\hline
CLIPStyler & VQ-GAN+CLIP & Diffusion+CLIP &Atlas baseline & Frames baseline \\
\hline&&&\\[-1em]
0.85 $\pm$ 0.12 & 0.86 $\pm$ 0.14  & 0.82 $\pm$ 0.11 & 0.73 $\pm$ 0.14 & 0.74 $\pm$ 0.15 \\
\hline
\end{tabular}
\end{center}
\caption{\emph{AMT surveys evaluation (see Sec.~\ref{sec:results})}. We compare to prominent (mask-free) image baselines (left), and demonstrate the effectiveness of our design choices in the video framework compared to alternatives (right). We report the percentage of judgments in our favor (mean, std). Our method outperforms all baselines.}\afterfigure
\label{table:AMT}
\end{table}

\subsection{Ablation Study}
\label{sec:ablation}
Fig.~\ref{fig:bootstrapping}(top) illustrates the effect of our relevancy-based bootstrapping (Sec.~\ref{sec:method_im}). As seen, this  component allows us to achieve accurate object mattes, which significantly improves the rough, inaccurate relevancy maps.

We ablate the different loss terms in our objective by qualitatively comparing our results when training with our full objective (Eq.~\ref{eq:total_loss}) and with a specific loss removed. The results are shown in Fig.~\ref{fig:bootstrapping}. 
As can be seen, without $\mathcal{L}_\mathsf{reg}$~(w/o sparsity), the output matte does not accurately capture the mango, resulting in a global color shift around it. Without $\mathcal{L}_\mathsf{structure}$~(w/o structure), the model outputs an image with the desired appearance  but fails to preserve the mango shape fully. Without $\mathcal{L}_\mathsf{screen}$~(w/o screen), the segmentation of the object is noisy (color bleeding from the mango), and the overall quality of the texture is degraded (see SM for additional illustration). Lastly, we consider a test-time optimization baseline by not using our internal dataset but rather inputting to $G_\theta$ the same input at each training step. As seen, this baseline results in lower-quality edits.
\begin{figure}[t!]
    \centering
    \includegraphics[width=\textwidth]{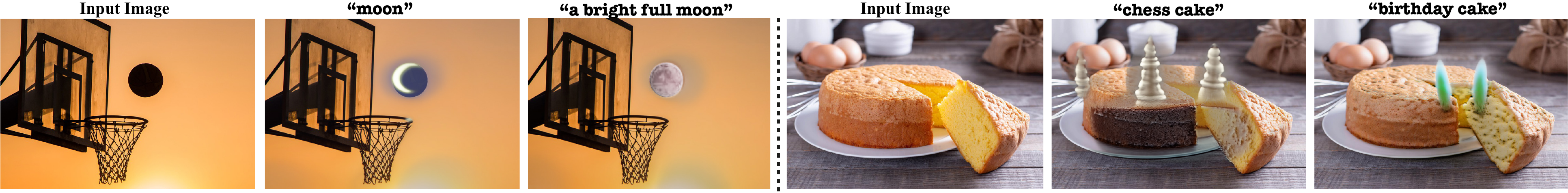}
    \vspace{-0.5cm}
    \caption{{\em Limitations.} CLIP often exhibit strong association between text and certain visual elements such as the shape of objects (e.g., ``moon'' with crescent shape), or additional new objects (e.g., ``birthday cake'' with candles). As our method is designed to edit existing objects, generating new ones may not lead to a visually pleasing result. However, often the desired edit can be achieved by using more specific text (left).} \afterfigure
    \label{fig:limitations_fig}
\end{figure}

\subsection{Limitations}
 We noticed that for some edits, CLIP exhibits a very strong bias towards a specific solution. For example, as seen in Fig.~\ref{fig:limitations_fig}, given an image of a cake,   the  text ``birthday cake''  is strongly associated with candles. Our method is not designed to significantly deviate from the input image layout and to  create new objects, and  generates unrealistic candles. Nevertheless, in many cases the desired edit can be achieved by using more specific text. For example,   the text ``moon'' guides the generation towards a crescent. By using the text ``a bright full moon''  we can steer the generation towards  a full moon  (Fig.~\ref{fig:limitations_fig} left). Finally, as acknowledged by prior works (e.g.,~\cite{text2mesh}), we also noticed that slightly different text prompts describing similar concepts may lead to slightly different flavors of edits. 

On the video side, our method assumes that the pre-trained NLA model accurately represents the original video. Thus, we are restricted to examples where NLA works well, as artifacts in the atlas representation can propagate to our edited video. An exciting avenue of future research may include fine-tuning the NLA representation jointly with our model.

\section{Conclusion}
We considered a new problem setting in the context of zero-shot text-guided editing: semantic, localized editing of existing objects within real-world images and videos. Addressing this task requires careful control of several aspects of the editing: the edit localization,  the preservation of the original content, and visual quality. We proposed to generate text-driven edit layers that allow us to tackle these challenges, without using a pre-trained generator in the loop. We further demonstrated how to adopt our image framework, with only minimal changes, to perform consistent text-guided video editing.  We believe that the key principles exhibited in the paper hold promise for leveraging large-scale multi-modal networks in tandem with an internal learning approach. 

\section{Acknowledgments}  We thank Kfir Aberman, Lior Yariv, Shai Bagon, and Narek Tumanayan for their insightful comments. We thank Narek Tumanayan for his help with the baselines comparison. This project received funding from the Israeli Science Foundation (grant 2303/20).

\clearpage
%
%
\bibliographystyle{splncs04}
\bibliography{egbib}

\appendix
\section{Implementation Details}\label{sec:implementation_details}
We provide implementation details for our architecture and training regime. 
\subsection{Generator Network Architecture}

We base our generator $G_\theta$ network on the \texttt{U-Net} architecture~\cite{ronneberger2015u}, with a 7-layer encoder and a symmetrical decoder. All layers comprise $3\!\times\!3$ Convolutional layers, followed by \texttt{BatchNorm}, and \texttt{LeakyReLU} activation. The intermediate channels dimensions is 128.
In each level of the encoder, we add an additional $1\!\times\!1$ Convolutional layer and concatenate the output features to the corresponding level of the decoder. Lastly, we add a $1\!\times\!1$ Convolutional layer followed by \texttt{Sigmoid} activation to get the final RGB output. 

\subsection{Internal Dataset (Sec. 3.1)} \label{sec:internal-dataset}
We apply data augmentations to the source image and target text {$(I_s$, $T)$} to create multiple \emph{internal examples} $\{(I^i_s, T^i)\}_{i=1}^N$. Specifically, at each training step, we apply a random set of image augmentations to $I_s$, and augment $T$ using a pre-defined set of text templates, as follows: 

\subsubsection{Image augmentations}
\begin{itemize}
  \item Random spatial crops: 0.85 and 0.95 of the image size in our image and video frameworks respectively. 
  \item Random scaling:  aspect ratio preserved scaling, of both spatial dimensions by a random factor, sampled uniformly from the range $[0.8,1.2]$.
  \item Random horizontal-flipping is applied with probability {p=0.5}.
  \item Random color jittering: we jitter the global brightness, contrast, saturation and hue of the image.
\end{itemize}
\subsubsection{Text augmentations and the target text prompt $T$}
 We compose $T$ with a random text template, sampled from of a pre-defined list of 14 templates. We designed our text-templates that does not change the semantics of the prompt, yet provide variability in the resulting CLIP embedding e.g.: 
{\small \begin{itemize}
  \item "photo of \{\}."
  \item "high quality photo of \{\}."
  \item "a photo of \{\}."
  \item "the photo of \{\}."
  \item "image of \{\}."
  \item "an image of \{\}."
  \item "high quality image of \{\}."
  \item "a high quality image of \{\}."
  \item "the \{\}."
  \item "a \{\}."
  \item "\{\}."
  \item "\{\}"
  \item "\{\}!"
  \item "\{\}..."
\end{itemize}}

At each step, one of the above templates is chosen at random and the target text prompt $T$ is plugged in to it and forms our augmented text. By default, our framework uses a single text prompt $T$, but can also support multiple input text prompts describing the same edit, which effectively serve as additional text augmentations (e.g., ``crochet swan'', and ``knitted swan'' can both be used to describe the same edit).

\subsection{Training Details}\label{sec:training_details}
We implement our framework in PyTorch \cite{NEURIPS2019PyTorch} (code will be made available). As described in Sec. 3, we leverage a pre-trained CLIP model ~\cite{clip} to establish our losses. We use the \texttt{ViT-B/32} pretrained model (12 layers, 32x32 patches),
downloaded from the  \href{https://github.com/openai/CLIP}{official implementation at GitHub}.
We optimize our full objective (Eq.~2, Sec.~3.1), with relative weights: $\lambda_g=1$, $\lambda_s=2$ (3 for videos), $\lambda_r=5\cdot 10^{-2}$, ($5\cdot 10^{-4}$ for videos) and $\gamma=2$. For bootstrapping, we set the relative weight to be 10, and for the image framework we anneal it linearly throughout the training.
We use the MADGRAD optimizer~\cite{defazio2021adaptivity} with an initial learning rate of $2.5\cdot 10^{-3}$, weight decay of 0.01 and momentum 0.9. We decay the learning rate with an exponential learning rate scheduler with $\text{gamma}=0.99$ ($\text{gamma}=0.999$ for videos), limiting the learning rate to be no less than $10^{-5}$. Each batch contains $(I^i_s, T^i)$ (see Sec. 3.1), the augmented source image and target text respectively. Every 75 iterations, we add {$\{I_s$, $T\}$} to the batch (i.e., do not apply augmentations). The output of $G_\theta$ is then resized down to $224$[px] maintaining aspect ratio and augmented (e.g., geometrical augmentations) before extracting CLIP features for establishing the losses. We enable feeding to CLIP arbitrary resolution images (i.e., non-square images) by interpolating the position embeddings (to match the size of spatial tokens of a the given image) using bicubic interpolation, similarly to \cite{dino}. 

Training on an input image of size $512\!\times\!512$ takes $\sim\!9$  minutes to train on a single GPU (NVIDIA RTX 6000) for a total of 1000 iterations.
Training on one video layer (foreground/background) of 70 frames with resolution $432 \times 768$ takes $\sim$60 minutes on a single GPU (NVIDIA RTX 8000) for a total of 3000 iterations.

\subsection{Video Framework}\label{sec:video_details}

We further elaborate on the framework's details described in Sec. 3.2 of the paper.

\myparagraph{Atlas Pre-processing.} Our framework works on a discretized atlas, which we obtain by rendering the atlas to a resolution of 2000$\times$2000 px. This is done as in \cite{kasten2021layered}, by querying the pre-trained atlas network in uniformly sampled UV locations.  The neural atlas representation is defined within the [-1,1] continuous space, yet the video content may not occupy the entire space. To focus only on the used atlas regions, we crop the atlas prior to training, by mapping all video locations to the atlas and taking their bounding box. Note that for foreground atlas, we map only the foreground pixels in each frame, i.e., pixels for which the foreground opacity is above 0.95;  the foreground/background opacity is estimated by the pre-trained neural atlas representation.

\myparagraph{Training.} As discussed in Sec.~3.2 in the paper, our generator is trained on atlas crops, yet our losses are applied to the resulting edited frames.   In each iteration, we crop the atlas by first sampling a video segment of 3 frames and mapping it to the atlas. Formally,  we sample a random frame $t$ and a random spatial crop size $(W,H)$ where its top left coordinate is at $(x, y)$. As a result we get a set of cropped (spatially and temporally) video locations:
{\small 
\begin{equation}
\begin{matrix}
\mathcal{V}=\{p=(x+j,y+i,t+m) & & & \text{   s.t.} & & & 0\leq j<W, & 0\leq i<H, & m\in\{-k, 0, k\} \}
\end{matrix}
\label{eq:sampled_video_locations}
\end{equation}} 
where $k=2$ is the offset between frames.

The video locations set $\mathcal{V}$ is then mapped to its corresponding UV atlas locations: $\mathcal{S}_{\mathcal{V}} = \mathbb{M}(\mathcal{V})$, where  $\mathbb{M}$ is a pre-trained mapping network. We define the atlas crop $I_{Ac}$ as the minimal crop in the atlas space that contains all the mapped UV locations:
{\small 
\begin{equation}
I_{Ac}=\left\{
\begin{matrix}
I_A[u,v] & & &s.t. & &  \text{min}(\mathcal{S_{\mathcal{V}}}.u) \leq u \leq \text{max}(\mathcal{S_{\mathcal{V}}}.u) \\
& & & & & \text{min}(\mathcal{S_{\mathcal{V}}}.v) \leq v \leq \text{max}(\mathcal{S_{\mathcal{V}}}.v),
\end{matrix}
\right\}
\label{eq:atlas_crop_def}
\end{equation}} 

We augment the atlas crop $I_{Ac}$ as well as the target text $T$, as described in Sec.~\ref{sec:internal-dataset} herein to generate an internal training dataset. To apply our losses, we map back the atlas edit layer to the original video segment and process the edited frames the same way as in the image framework: resizing, applying CLIP augmentations, and applying the final loss function of Eq.~2 in Sec. 3.1 in the paper. To enrich the data, we also include one of the sampled frame crops as a direct input to $G$ and apply the losses directly on the output (as in the image case).   Similarly to the image framework, every 75 iterations we additionally pass the pair $\{I_A, T\}$, where $I_A$ is the entire atlas (without augmentations, and without mapping back to frames). For the background atlas, we first downscale it by three due to memory limitations.

\myparagraph{Inference.} As described in Sec.~3.2, at inference time, the entire atlas $I_A$ is fed into $G_\theta$ results in $\mathcal{E}_A$. The edit is mapped and combined with the original frames using the process that is described in \cite{kasten2021layered}(Sec. 3.4, Eq. (15),(16)). Note that our generator operates on a single atlas. To produce foreground and background edits, we train two separate generators for each atlas.

\end{document}